\documentclass[letterpaper]{article}
\usepackage{aaai24}
\usepackage{times}
\usepackage{helvet}
\usepackage{courier}
\usepackage[hyphens]{url}
\usepackage[pdftex]{graphicx}
\usepackage{natbib}
\usepackage{caption}
\nocopyright

\title{Meta-Learning for Neural Network-based Temporal Point Processes}
\author{
    Yoshiaki Takimoto\textsuperscript{\textrm{1}},
    Yusuke Tanaka\textsuperscript{\textrm{2}},
    Tomoharu Iwata\textsuperscript{\textrm{2}},
    Maya Okawa\textsuperscript{\textrm{1}}\footnote{Current address: Physics \& Informatics Laboratories, NTT Research, Inc.},\\
    Hideaki Kim\textsuperscript{\textrm{1}},
    Hiroyuki Toda\textsuperscript{\textrm{1}}\footnote{Current address: School of Data Science, Yokohama City University},
    Takeshi Kurashima\textsuperscript{\textrm{1}}
}
\affiliations{
    \textsuperscript{\rm 1}NTT Human Informatics Laboratories\\
    \textsuperscript{\rm 2}NTT Communication Science Laboratories\\
    yoshiaki.takimoto@ntt.com\\
    ysk.tanaka@ntt.com\\
    tomoharu.iwata@ntt.com\\
    maya.okawa@ntt-research.com\\
    hideaki.kin@ntt.com\\
    hirotoda@acm.org\\
    takeshi.kurashima@ntt.com
}

\usepackage{comment}
\usepackage{multirow}

\usepackage[subrefformat=simple]{subcaption}

\usepackage{amsthm}
\newtheorem{theorem}{Theorem}

\usepackage{amsmath}
\usepackage{amssymb}
\usepackage{amsfonts}
\usepackage{mathtools}
\usepackage{bm}
\usepackage{siunitx}
\providecommand{\abs}[1]{\left\lvert#1\right\rvert}

\usepackage{placeins}

\usepackage{array}
\usepackage{hhline}
\usepackage{tablefootnote}
\usepackage{booktabs}
\usepackage{dcolumn}

\usepackage[linesnumbered,ruled,vlined]{algorithm2e}
\SetAlgorithmName{Algorithm}{}{}

\usepackage[noabbrev,capitalize]{cleveref}

\crefname{equation}{}{}
\crefformat{footnote}{#2\footnotemark[#1]#3}
\usepackage{refcount}

\newcommand{\cycle}{\mathrm{peri}}
\newcommand{\noncycle}{\mathrm{aperi}}

\begin{document}

\maketitle

\begin{abstract}
    Human activities generate various event sequences such as taxi trip records, bike-sharing pick-ups, crime occurrence, and infectious disease transmission.
    The point process is widely used in many applications to predict such events related to human activities.
    However, point processes present two problems in predicting events related to human activities.
    First, recent high-performance point process models require the input of sufficient numbers of events collected over a long period (i.e., long sequences) for training, which are often unavailable in realistic situations.
    Second, the long-term predictions required in real-world applications are difficult.
    To tackle these problems, we propose a novel meta-learning approach for periodicity-aware prediction of future events given short sequences.
    The proposed method first embeds short sequences into hidden representations (i.e., task representations) via recurrent neural networks for creating predictions from short sequences.
    It then models the intensity of the point process by monotonic neural networks (MNNs), with the input being the task representations.
    We transfer the prior knowledge learned from related tasks and can improve event prediction given short sequences of target tasks.
    We design the MNNs to explicitly take temporal periodic patterns into account, contributing to improved long-term prediction performance.
    Experiments on multiple real-world datasets demonstrate that the proposed method has higher prediction performance than existing alternatives.
\end{abstract}

\section{Introduction\label{sec:intro}}
Human activities generate various event sequences such as taxi trip records \cite{pang2017discovering}, bike-sharing pick-ups \cite{gervini2019exploring}, crime occurrence \cite{mohler2011self}, and infectious disease transmissions \cite{choi2015constructing}.
Such event sequences carry rich information with timestamps in terms of continuous time and event attributes (e.g., location) indicating when and what events occurred.
Modeling event sequences and predicting future events benefit many applications, including traffic management, predictive policing, and epidemic control.
For example, if one could accurately predict when and where bike-sharing pick-ups will occur, the bike-share companies could efficiently allocate their bikes to the appropriate places in advance \cite{de2016bike}.

In the task of predicting events associated with human activities, point processes have emerged as a prevalent tool \cite{kim2014tracking,kim2017read,bacry2015hawkes,mohler2011self}.
Point process models are characterized by an intensity function that determines the probability of an event being observed over time.
Recent studies \cite{shchur2021neural} examined improving the expressiveness of these intensity functions by combining them with deep neural networks.
Notably, \citet{du2016recurrent} used a recurrent neural network (RNN) in modeling the intensity of point processes.
Building upon this, \citet{omi2019fully} extended this approach by integrating RNN with a monotonic neural network (MNN) \cite{sill1997monotonic} to further enhance the flexibility of intensity.

However, in practical applications where point processes are used to predict events related to human activities, there remain two main challenges.
First, predictive performance decreases when \emph{the observation period of events is short}.
Although these neural network-based point process methods show superior event prediction performance, they need long sequences of events for model training, which are often unavailable in realistic situations.
For example, bike-sharing systems often build new bike stations to meet the increasing demand.
The newly opened stations provide only a limited number of events just after being opened (i.e., short sequences).
As a result, prediction performance is poor and makes it difficult to allocate bikes ahead of time properly.
Moreover, in certain situations, unanticipated behavioral changes, such as sudden increases in taxi demand due to specific things (e.g., construction projects and social gatherings), require prediction flexibility as only recent, short-term data is available.
Meta-learning offers a promising solution to such issues, especially when we have only a short observation period for certain tasks (e.g., new bike stations), while longer periods are available for others (e.g., existing bike stations).
In this vein, \citet{xie2019meta} introduced HARMLESS, which integrated model-agnostic meta-learning (MAML) and Hawkes processes.
However, MAML demands adaptation through the gradient method, incurring significant memory overheads.
Moreover, to prevent overfitting, it necessitates the use of simpler models like the Hawkes process, consequently limiting its expressive power.

Second, \emph{the prediction accuracy declines for long-term predictions}, such as those spanning over a week.
This presents practical issues for our intended applications including security planning.
Implementing the prediction results, which includes managing manpower availability and formulating plans, often takes an extended period, ranging from several days to a week.
\citet{kovacsWeightedFairResource2011,pozdnoukhov2010exploratory,malmgren2009characterizing} have incorporated periodicity into the point process.
However, their intensity functions' expressive powers tend to be constrained because they depend on parametric assumptions.

To tackle these challenges, we propose a novel meta-learning approach for periodicity-aware prediction of future events given short sequences.
In our framework, we first derive task representations from the short sequences by embedding them via RNN.
We then describe the intensity of point processes based on the task representations and use the intensity to predict future events.
The model is trained using test likelihood for the long-term prediction and following the framework of the neural processes \cite{garnelo2018neural}, a meta-learning method.
By integrating RNN, informed by meta-learning, into our formulation, we can fully exploit the temporal information from the \emph{short sequences}, thus addressing the first challenge.
We design the intensity using an MNN that takes a timestamp and task representations as inputs.
This design allows our model to learn flexible representations of the event sequences while keeping learning tractable; that is, the model parameter can be estimated based on the exact likelihood.
Moreover, we further extend an MNN-based intensity by combining it with a periodic function of time.
By integrating the extended MNN into our framework, we can automatically learn daily or weekly \emph{periodic patterns} in the event sequences, thus addressing the second challenge.
The MNN is further extended to incorporate urban contexts such as land use and community assets.

We carried out experiments using three real-world datasets to validate the efficacy of our proposed method for long-term event prediction based on short sequences.
The experiments reveal that our method outperforms the four existing methods in settings with limited data.
Additionally, the findings indicate that our approach is adept at identifying periodic variations and efficiently considers urban contexts.

The main contributions of this paper are as follows.
\begin{enumerate}
      \item We propose a meta-learning framework for point process modeling with short sequences of events.
            This framework allows new tasks with short sequences to effectively predict future events by leveraging the knowledge of tasks with long sequences.
            It also has the advantage of saving memory because it does not require gradient-based adaptation to new tasks.
      \item We extend MNN by combining it with a periodic function of time as well as incorporating urban contexts while keeping learning tractable.
            This yields superior performance in the long-term prediction of future events.
            To the best of our knowledge, this is the first method to explicitly incorporate prior knowledge of periodicity into meta-learning of point processes.

\end{enumerate}

\section{Related Work\label{sec:related}}
\subsection{Point Process\label{subsec:pp}}
Point processes have been widely used to model event sequences in various fields \cite{yan2019recent,shchur2021neural} such as taxi usage records \cite{pang2017discovering}, bike-sharing pick-ups \cite{gervini2019exploring}, software reliability engineering \cite{huang2003unified}, infectious disease transmission \cite{choi2015constructing}, and financial transactions \cite{bacry2015hawkes}.
In point processes, intensity functions represent the probability of an event occurring.
Intensity functions are learned by maximum likelihood estimation.
The common types of intensity functions are constant intensity functions (i.e., homogeneous Poisson processes), time-varying intensity functions (i.e., inhomogeneous Poisson processes) \cite{pointprocess2005}, and intensity functions whose values change based on the history of event occurrences (e.g., Hawkes process \cite{hawkes1971point}).
Researchers have manually designed intensity functions, for example, by incorporating periodicity into inhomogeneous Poisson processes \cite{kovacsWeightedFairResource2011,pozdnoukhov2010exploratory,malmgren2009characterizing}.
Their expressive powers tend to be constrained because they depend on parametric assumptions.
In order to improve the expressiveness of the intensity function, deep learning methods have been proposed in recent years \cite{omi2019fully,mei2016neural,du2016recurrent,gu2021attentive,menon2017predicting}.
For example, \citet{menon2017predicting} combine an inhomogeneous Poisson process with a neural network to learn from bus usage records and then predict future usage.
In addition, more recent works \cite{omi2019fully,mei2016neural,du2016recurrent,gu2021attentive} combine Hawkes processes with neural networks such as RNNs and attention mechanisms.
However, they focus on predicting the timestamp of the next event based on the history of events; namely, long-term prediction is difficult.
Also, they assume that long sequences of the prediction target are available for training.
They do not consider the situation where only short sequences of the prediction target are available.

\subsection{Meta-Learning\label{subsec:fsl}}
Meta-learning methods have shown promising results in a variety of few-shot learning settings, including face recognition \cite{guo2020learning}, object detection \cite{wang2019meta}, robotics imitation learning \cite{finn2017one}, density estimation \cite{reed2017few}, and generative modeling \cite{rezende2016one,bornschein2017variational}.
Several works have proposed meta-learning approaches for sequential data \cite{iwata2020few,ribeiro2018transfer,talagala2018meta,prudencio2004meta,lemke2010meta,hooshmand2019energy,ali2018cross}; however, they cannot be used for event sequences over continuous time.
Very few methods present a gradient-based meta-learning approach for predicting future events from short sequences.
\citet{xie2019meta} combine MAML \cite{finn2017model} and Hawkes processes \cite{hawkes1971point}.
As their intensity function depends on parametric assumptions such as the exponential function, its expressive power is constrained.
Since MAML incurs high computation costs due to its use of higher-order derivatives, this method cannot easily be scaled to large models (e.g., neural networks) built to determine the intensity function.
Moreover, it does not incorporate external factors (e.g., temporal periodicity and urban contexts) that affect human activities.
Our proposed method adopts another meta-learning approach based on black-box adaptation \cite{garnelo2018neural}, i.e., without second-order derivatives.
In our approach, short sequences are embedded into internal representations via RNN and used as the inputs of the meta-learner.
Unlike MAML-based methods, the proposed method does not require adaptation using the gradient method and supports external information (i.e., temporal periodicity and urban contexts).

\section{Preliminary\label{sec:preliminary}}
Our proposed method is based on temporal point processes \cite{pointprocess2005},
which are widely used for modeling and predicting
discrete time event occurrences on continuous timelines.
We briefly describe the temporal point process framework below.

Let $t\in\mathbb{R}_{\geq 0}$ be the time of event occurrence.
In the temporal point process, event occurrence is assumed to be governed by the \emph{intensity function} $\lambda(t)\geq 0$ defined as
\begin{equation}
  \lambda(t)
  =\lim_{\Delta\rightarrow0}
  \frac{p(\mathrm{one\ event\ occurs\ in}\ [t,t+\Delta])}{\Delta},
\end{equation}
where $\Delta$ is a time interval.
$\lambda(t)$ is the instantaneous probability of event occurrence at time $t$;
one can model the dynamics of event occurrence by designing $\lambda(t)$ to suit the application.

Assume that we obtain the event sequence
$X_T=\{t_1,\dots,t_N\},(0\leq t_1\leq \dots \leq t_N \leq T)$,
where $T$ is the observation period, $t_n$ is the $n$-th event, and $N$ is the number of events observed within that period.
The likelihood of $X_T$ is given by
\begin{equation}\label{eq:pre_likelihood}
  p(X_T) = \prod_{t_i \in X_T}\exp\left[-\int_{0}^{T}\lambda(t)dt \right]\lambda(t_i) .
\end{equation}
The intensity function can be estimated by maximizing the logarithm of $p(X_T)$.
One can also predict event occurrences using the estimated intensity function.

\section{Proposed Method\label{sec:propose}}
\subsection{Problem Setting\label{subsec:task}}
We first define the word \emph{task}.
It represents one problem of estimating the future intensity function from a short event sequence.
Note that the problem setting described in \cref{sec:preliminary} corresponds to a \emph{single task}.
In contrast, this work handles \emph{multiple tasks}.
For example, in the bike-sharing system, a task is modeling the dynamics of bike pick-ups occurring at a bike station, and we consider bike pick-ups occurring at multiple bike stations.
Our goal is to predict future events of an \emph{unseen target task} from just a short event sequence by utilizing a large amount of data from multiple tasks.
This problem is significant in various applications; for instance, one can predict future usage of a \emph{new} bike station from just a short sequence in its early stages, which is useful for optimizing the allocation of bikes to stations.

We describe below the mathematical notations of our problem setting.
Let $M$ denote a set of tasks, $m\in M$ denote a task, and $g \in\mathbb{R}^{K_g}$ be the urban contexts such as land use and community assets.
In the training phase, we assume that a set of paired event sequences and urban contexts $\mathcal{D}=\{(X_{T^\mathrm{e}}^{(m)}, g^{(m)})\}_{m\in M}$ is available, where $X_{T^\mathrm{e}}^{(m)}$ is event sequence $X_{T^\mathrm{e}}$ of the task $m$.
In the test phase, given the short sequence $X_{T^\mathrm{c}}^{(m^*)}$ up to current time $T^\mathrm{c}$ and urban contexts $g^{(m^*)}$ in the new task, $m^*$, we wish to predict the intensity function $\lambda^{(m^*)}(t)$ for task $m^*$ in the query time interval $[T^\mathrm{c}, T^\mathrm{e}]$.
It should be noted that $T^\mathrm{c}\ll T^\mathrm{e}$ and $m^* \notin M$.
Note that we do not update the prediction based on future events $\{t_i^{(m^*)} \mid T^{\mathrm{c}} < t_i^{(m^*)} \leq T^{\mathrm{e}}\}$, but only one prediction at $T^\mathrm{c}$.
Although we assume that the observation period for all tasks is $[0,T^{\mathrm{e}}]$ and the query time interval $(T^\mathrm{c}, T^\mathrm{e}]$ for simplicity, our model is easily extended to handle event sequences of different lengths.

\begin{figure}[t]
    \includegraphics[width=\linewidth]{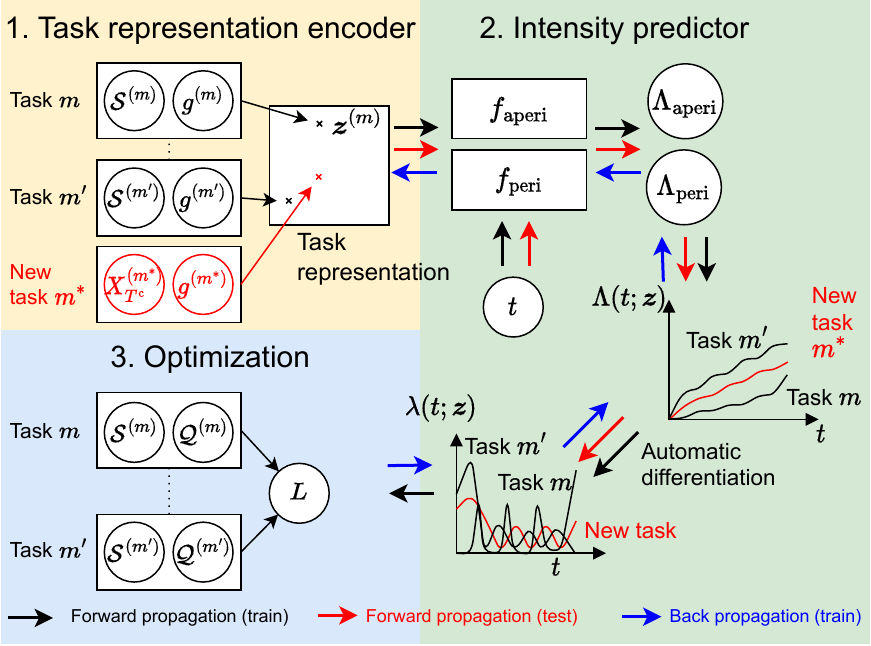}
    \caption{Our model.\label{fig:model}
        \textmd{
            \textbf{Training phase:} First, task representation $\bm{z}$ is inferred from each support set $\mathcal{S}$ and urban contexts $g$.
            Next, task-specific intensity function $\lambda(t;\bm{z})$ is estimated from task representation $\bm{z}$ at future time $t$.
            $\lambda(t;\bm{z})$ is the sum of periodic intensity function $\lambda_{\cycle}(t;\bm{z})$ and aperiodic intensity function $\lambda_{\noncycle}(t;\bm{z})$.
            The integral of $\lambda_{\cycle}(t;\bm{z})$ and $\lambda_{\noncycle}(t;\bm{z})$ i.e., $\Lambda_{\cycle}(t;\bm{z})$ and $\Lambda_{\noncycle}(t;\bm{z})$ are modeled by MNNs $f_\cycle(t, \bm{z})$ and $f_\noncycle(t, \bm{z})$.
            Then, loss $L$ is calculated from the task-specific intensity function and learned by backpropagation.
            \textbf{Test phase:}  First, task representation $\bm{z}$ is inferred from support set $X_{T^\mathrm{c}}^{(m^*)}$ and urban contexts $g^{(m)}$ of the new task.
            Then, the intensity function is estimated as in the training phase.
        }
    }
\end{figure}
\subsection{Overview of Model\label{subsec:model}}
To solve the problem of predicting future events in the new task from a short sequence, we propose a novel meta-learning framework for temporal point processes.
We show a schematic diagram of the proposed framework in \cref{fig:model}.
We design the training procedure for simulating the test phase to be similar to the episodic training framework \cite{finn2017model}.
Training data $X^{(m)}_{T^\mathrm{e}}$ in each task are divided into two disjoint subsets of \emph{support set} $\mathcal{S}^{(m)}=X^{(m)}_{T^\mathrm{c}}=\{t_i^{(m)}\mid 0 \leq t_i^{(m)} \leq T^{\mathrm{c}}\}$ and \emph{query set} $\mathcal{Q}^{(m)}=\{t_i^{(m)}\mid T^{\mathrm{c}} < t_i^{(m)} \leq T^{\mathrm{e}}\}$.
Note that the support set contains the events in the short period $[0,T^{\mathrm{c}}]$ captured in the initial stage.
As one can see from~\cref{fig:model},
support sets $\{\mathcal{S}^{(m)}\}$ and urban contexts $\{g^{(m)}\}$ are used for inferring the intensity functions; the loss function is calculated using query sets $\{\mathcal{Q}^{(m)}\}$.
This procedure allows the model to be trained so that it can predict future events of the new task even if the initial sequence of that task is short.
Moreover, our framework has two important components, i.e.,
\emph{task representation encoder} and \emph{intensity predictor}.
The task representation encoder embeds a sequence and urban contexts in a task into the same latent space, called task representation, which allows for estimation of the relationships among tasks.
The intensity predictor is a neural network-based intensity function that incorporates explicit modeling of periodic behaviors.
This formulation is advantageous because it can model flexible intensity functions while accounting for the periodicity in human activities.
We specify the architectures in \cref{subsec:representation,subsec:intensity} and the algorithm for meta-learning in \cref{subsec:meta-learning}.

\subsection{Task Representation Encoder\label{subsec:representation}}
To model the relationships among tasks, we introduce task representation $\bm{z}\in\mathbb{R}^{K_{\mathrm{z}}}$, which is calculated from support set $\mathcal{S}^{(m)}$ and urban contexts $g^{(m)}$ by neural networks.
First, we embed the support set to obtain support representation $\bm{z}_\mathcal{S}\in\mathbb{R}^{K_\mathcal{S}}$.
Then we incorporate urban contexts into the support representation to obtain task representation $\bm{z}$.

\textbf{Support Representation.}
Event sequences are typically interrelated; for example, popular spots attract many people.
In order to capture such interrelationships between event occurrences, we use an RNN as the encoder.
The representation of task $m$ is given by
\begin{equation}\label{eq:rnn}
    \bm{z}^{(m)}_\mathcal{S} =
    \frac{1}{\abs{\mathcal{S}^{(m)}}}
    \sum_{n=1}^{\abs{\mathcal{S}^{(m)}}}
    \mathrm{RNN}\left(t_{n}^{(m)},t_{n}^{(m)}-t_{n-1}^{(m)}\right),
\end{equation}
where the inputs of the RNN are
timestamp $t_n^{(m)}$ of the $n$-th event and time interval $t_n^{(m)}-t_{n-1}^{(m)}$.
Here $\abs{\cdot}$ represents the number of elements in a set, and $t_0=0$ is assumed.

\textbf{Task Representation.}
Urban contexts are also related to event occurrences; for example, residential regions have more morning and evening trips due to commuters, while commercial regions have more trips by shoppers during the day.
In order to capture such relations, we incorporate urban contexts in the support representation.
The representation of task $m$ is given by
\begin{equation}\label{eq:rnn_external}
    \bm{z}^{(m)} =
    \mathrm{FNN}\left(\bm{z}^{(m)}_\mathcal{S}, g^{(m)}\right),
\end{equation}
where $\mathrm{FNN}:\mathbb{R}^{K_\mathcal{S}+K_\mathrm{g}}\to\mathbb{R}^{K_{\mathrm{z}}}$ is a fully-connected neural network.

It should be noted that the RNN in~\cref{eq:rnn} and the FNN in~\cref{eq:rnn_external} are shared among all tasks.
The RNN and FNN make it possible to properly capture the initial patterns of the respective sequences.
Therefore, the proposed method is expected to effectively predict the intensity functions in the future by discerning the differences in the initial patterns of short sequences depending on urban contexts.

\subsection{Intensity Predictor\label{subsec:intensity}}
In this section, we describe the design of intensity function $\lambda(t;\bm{z}^{(m)})$ for task $m$.
Here, we assume that the intensity depends on task representation $\bm{z}^{(m)}$ defined by~\cref{eq:rnn_external}.
In human activity data, the intensity often demonstrates periodic dynamics; thus, we model the intensity function as the sum of \emph{periodic} and \emph{aperiodic} terms, as follows:
\begin{equation}\label{eq:intenisty_definition}
    \lambda\left(t;\bm{z}^{(m)}\right) =
    \lambda_\cycle\left(t;\bm{z}^{(m)}\right)
    +\lambda_\noncycle\left(t;\bm{z}^{(m)}\right),
\end{equation}
where $\lambda_\cycle(t;\bm{z}^{(m)})$ and $\lambda_\noncycle(t;\bm{z}^{(m)})$
are the periodic and the aperiodic intensity functions, respectively.
Neural networks are promising for modeling flexible intensity functions; however, if $\lambda_\cycle(t;\bm{z}^{(m)})$ and $\lambda_\noncycle(t;\bm{z}^{(m)})$
are modeled directly by neural networks, the integral of the intensity contained in the likelihood (See~\cref{eq:pre_likelihood}) is intractable.
The adoption of the common solution, numerical approximation, results in poor fitting accuracy and high computation cost.
To avoid this difficulty, we model the integral of the intensity function over $[0,t]$ using neural networks, as in~\cite{omi2019fully}.
The integral of~\cref{eq:intenisty_definition} is given by
\begin{align}\label{eq:cumulative_definition}
    \Lambda\left(t;\bm{z}^{(m)}\right)
     & := \int_0^t \lambda\left(u;\bm{z}^{(m)}\right) du
    \nonumber                                            \\
     & = \Lambda_\cycle\left(t;\bm{z}^{(m)}\right)
    + \Lambda_\noncycle\left(t;\bm{z}^{(m)}\right),
\end{align}
where $\Lambda_\cycle(t;\bm{z}^{(m)})$ and
$\Lambda_\noncycle(t;\bm{z}^{(m)})$
are the cumulative functions of $\lambda_\cycle(t;\bm{z}^{(m)})$
and $\lambda_\noncycle(t;\bm{z}^{(m)})$, respectively.
Since the cumulative functions are modeled by neural networks, we can obtain the corresponding intensities by differentiating the cumulative functions with respect to time $t$ via automatic differentiation.
Details of the periodic and aperiodic terms are described in the following paragraph.

\textbf{Periodic term.}
We design the periodic intensity function to satisfy the periodic constraint
$\lambda_\cycle(t;\bm{z}^{(m)}) = \lambda_\cycle(t+\tau;\bm{z}^{(m)})$, where $\tau$ is the period of human activity, such as a week or a day, as indicated by prior knowledge.
If more than one period exists, multiple $\tau$ can be prepared and divided into multiple periodic terms, such as $\lambda_\cycle(t;\bm{z}^{(m)}) = \lambda_{\cycle_1}(t;\bm{z}^{(m)}) + \lambda_{\cycle_2}(t;\bm{z}^{(m)})$.
Since the intensity function is non-negative by definition, its cumulative function is a monotonically increasing function to $t$; thus, we model the periodic term $\Lambda_\cycle(t; \bm{z}^{(m)})$ in~\cref{eq:cumulative_definition} using MNNs represented by $f_\cycle(t, \bm{z}^{(m)}):\mathbb{R}^{K_{\mathrm{z}}+1}\to\mathbb{R}$.
To ensure the output of the MNN is monotonically increasing with respect to the input, weights are restricted to non-negative values and the activation function is set to a monotonically increasing function, such as tanh and softplus.
$\Lambda_\cycle(t; \bm{z}^{(m)})$ is given by
\begin{align}\label{eq:cycle_dif_by_mnn}
    \MoveEqLeft \Lambda_\cycle\left(t;\bm{z}^{(m)}\right)                        \nonumber                                                     \\
     & = s\left(f_\cycle\left(t',\bm{z}^{(m)}\right) - f_\cycle\left(0,\bm{z}^{(m)}\right)\right)
    \nonumber                                                                                                                                  \\
     & + s\left\lfloor\frac{t}{\tau}\right\rfloor \left( f_\cycle\left(\tau,\bm{z}^{(m)}\right) - f_\cycle\left(0,\bm{z}^{(m)}\right) \right),
\end{align}
where $t'=t-\tau \lfloor\frac{t}{\tau}\rfloor$ is the phase of the periodic intensity function, and $\lfloor \cdot \rfloor$ is the floor function.
$s\in\mathbb{R}_{> 0}$ in~\cref{eq:cycle_dif_by_mnn} is a scale parameter determined from the statistics of training data $\mathcal{D}$.
It is used to adjust the value range of the intensity function like normalization.
In our implementation, $s$ is set to the maximum number of events over the query window as $s = \max_m \abs{\mathcal{Q}^{(m)}}$.
The derivative of $\Lambda_\cycle(t;{\bm{z}}^{(m)})$ i.e., $\lambda_\cycle(t;{\bm{z}}^{(m)})$, is the periodic function of period $\tau$ (\cref{theorem:periodicity}).
The reason for subtracting $f_\cycle(0,\bm{z}^{(m)})$ is to set $\Lambda_\cycle(0; \bm{z}^{(m)}) = \int_0^0\lambda_\cycle(0; \bm{z}^{(m)})=0$.
Details of \cref{theorem:periodicity} and proof are given in \cref{sec:proof}.

\textbf{Aperiodic term.}
This term is a black-box intensity function defined by neural networks.
For the same reasons of periodic term, aperiodic term $\Lambda_\noncycle(t;\bm{z}^{(m)})$ is also modeled by another MNN, as follows:
\begin{equation}\label{eq:noncycle_dif_by_mnn}
    \Lambda_\noncycle(t;\bm{z}^{(m)}) = s\left( f_\noncycle\left(t,\bm{z}^{(m)}\right)-f_\noncycle\left(0,\bm{z}^{(m)}\right)\right),
\end{equation}
where $f_\noncycle(t,\bm{z}^{(m)}):\mathbb{R}^{K_{\mathrm{z}}+1}\to\mathbb{R}$ is the MNN for the aperiodic term.

\subsection{Meta-Learning Algorithm\label{subsec:meta-learning}}
\begin{algorithm}[t]
    \DontPrintSemicolon
    \KwIn{Event datasets $\mathcal{D}$, query time interval $[T^\mathrm{c},T^\mathrm{e}]$}
    \KwOut{Trained neural network parameters $\theta$}
    Determine $s$ from $\mathcal{D}$\;
    \While{\textrm{not done}}{
        Randomly sample task $m$ from $M$\;
        Divide $X^{(m)}_{T^\mathrm{e}}$ into support set $\mathcal{S}$ and query set $\mathcal{Q}$\;
        Calculate task representation $\bm{z}$ by \cref{eq:rnn_external}\;
        Calculate loss by \cref{eq:loss}, and its gradients\;
        Update parameters $\theta$ using loss and its gradients
    }
    \caption{Training procedure of our model\label{algo:training}}
\end{algorithm}
Let $\theta$ be the parameters of the proposed neural networks (i.e., RNN, FNN, and MNNs).
The parameterization by $\theta$ is indicated by a subscript, e.g., the left side of \cref{eq:cumulative_definition} becomes $\Lambda_\theta(t;\bm{z}^{(m)})$.
We aim to estimate parameters $\theta$ based on the episodic training framework~\cite{finn2017model}.
The optimization problem to be solved is given by
\begin{equation}\label{eq:object}
    \hat{\theta} =
    \arg \min_{\theta}
    \mathbb{E}_{m\sim M}
    \left[
    \mathbb{E}_{(\mathcal{S},\mathcal{Q})
    \sim X^{(m)}_{T^\mathrm{e}}}[L_\theta(\mathcal{S},\mathcal{S}\cup\mathcal{Q})]
    \right],
\end{equation}
where $\mathbb{E}[\cdot]$ is the expectation.
Note that superscript $m$ is omitted from $\mathcal{S}^{(m)}$ and $\mathcal{Q}^{(m)}$.
The objective function $L_\theta(\mathcal{S},\mathcal{S}\cup \mathcal{Q})$ in~\cref{eq:object} is the following negative log-likelihood,
\begin{equation}\label{eq:loss}
    L_\theta(\mathcal{S},\mathcal{S} \cup \mathcal{Q})
    = -\sum_{t \in \mathcal{S} \cup\mathcal{Q}}
    \log \frac{\partial \Lambda_\theta(t;\bm{z})}{\partial t}
    + \Lambda_\theta(T^{\mathrm{e}};\bm{z}),
\end{equation}
where the first argument, $\mathcal{S}$, is for calculating task representation, and the second argument, $\mathcal{S}\cup\mathcal{Q}$, is the events to be optimized.
We include not only the query set but also the support set in the optimization process to stabilize the estimation around $T^\mathrm{c}$.
The second term is the cumulative function~\cref{eq:cumulative_definition} parameterized by neural networks; the first term can be obtained using automatic differentiation.
One observes that the objective function~\cref{eq:loss} can be obtained efficiently by avoiding the intractable integral in~\cref{eq:pre_likelihood}.
We show the training procedure in~\cref{algo:training}.

\section{Experiments\label{sec:experiments}}
\subsection{Datasets\label{subsec:data}}
Our experiments used three different real-world event datasets, Bikeshare dataset, Taxi dataset, and Crime dataset, as well as urban context data as described below.

\textbf{Bikeshare dataset.}
The Bikeshare dataset is the historical trip data collected by the NYC Citi Bike system\footnote{\url{https://ride.citibikenyc.com/system-data}}.
This data includes 114,193,076 bike trips captured by the 1,491 bike stations in New York City (NYC) over an eight-year period (July 1, 2013 - January 31, 2021).
It comprises a timestamped record of every trip with bike pick-up time and rental station.
We used 539 stations open before March 2020, and the stations opening between April and July 2020 for training and validation, respectively.
76 stations newly opened after August 2020 were selected to test the model's prediction performance.
We set the observation start time $t_0$ to 0:00 AM. on the day after the first use of the station.
We set $T^\mathrm{c}= 12$ hours and $T^\mathrm{e}= 7$ days.
We excluded the bike stations that were opened in the last week of each data period to avoid overlapping periods.
Also, we excluded the stations with less than five events in the support set to better support RNN input and MAML's adaptation.

\textbf{Taxi dataset \textmd{\cite{TaxiData}}.}
The Taxi data consists of 77,399,896 taxi trips in NYC.
The data period spans January 1, 2014 to December 31, 2014.
Each trip is associated with taxi pick-up time and location (i.e., GeoHash\footnote{\url{http://geohash.org/site/tips.html}} region).
We extracted 970, 276, and 328 Geohash level 7 regions for training, validation, and test, respectively.
We then extracted three consecutive days (from 5:00 AM to 5:00 AM three days later) from each region from January through May for the training, June for the validation, and July or later for the test to make up the event sequences, respectively.
We set $T^\mathrm{c}= 7$ hours and $T^\mathrm{e}= 3$ days.
We excluded regions with less than five events.

\textbf{Crime dataset\textmd{\footnote{\url{https://www.nyc.gov/site/nypd/stats/crime-statistics/citywide-crime-stats.page}}}.}
The crime dataset consists of 8,357,267 records in NYC.
Each record is associated with the time and location of the complaint.
We extracted 313, 83, and 80 Geohash level 6 regions for training, validation, and test, respectively.
We then extracted 21 consecutive days (from 5:00 AM to 5:00 AM 21 days later) from each region from January through June for the training, from July through September for the validation, and October or later for the test to make up the event sequences, respectively.
We set $T^\mathrm{c}= 36$ hours and $T^\mathrm{e}= 21$ days.
We excluded regions with less than five events.

We further collected the corresponding urban contexts for each bike station and GeoHash region.
The urban contexts used for all datasets were land use, community assets, number of crimes\footnote{For the Crime dataset, this data is excluded.}, street cleanliness, and mean commute time to work, which were downloaded from Community District Profiles\footnote{\url{https://communityprofiles.planning.nyc.gov/}\label{foot:NYC}}.
We refer to Community District via the latitude and longitude of the bike station for Bikeshare or the center of the region.
Statistics and urban contexts of the dataset are shown in \cref{sec:dataset}.

\subsection{Comparison Methods\label{subsec:VSmethods}}
We compared the proposed method with the following existing methods, further details can be found in \cref{subsec:VSmethods_describe}.

\textbf{HPP} (Homogeneous Poisson process) \cite{pointprocess2005} is a model with a constant intensity function.
The intensity function is determined based on the support set of each task.

\textbf{NNIPP} (Neural Network Inhomogeneous Poisson process)  is a simple neural-network-based inhomogeneous Poisson process model.
The inhomogeneous Poisson process \cite{pointprocess2005} has an intensity function that varies with time.
The network is the same as the aperiodic term of the proposed method.
In training and prediction, we regard all event data as the same task.

\textbf{NM} (NNIPP with Model-agnostic meta-learning) is a model that applies MAML \cite{finn2017model} to NNIPP.
MAML, a gradient-based approach, performs double-loop training to achieve meta-learning.
In the inner loop, parameters are updated several times to adapt to the new task.
The number of inner loops, denoted as $l$ in this paper, varied from 1 to 4.

\textbf{HARMLESS} \cite{xie2019meta} is a model that applies MAML to Hawkes process.
HARMLESS has multiple exponential kernels, the number of which is from 1 to 4.

\subsection{Experimental Setup\label{subsec:setup}}
\textbf{Evaluation Metrics.}
We used two metrics to evaluate the accuracy of event prediction.
The first metric is the negative test log-likelihood (NLL), a version of \cref{eq:loss} with only the query set as the test data (i.e., $L(\mathcal{S}^{(m^*)},\mathcal{Q}^{(m^*)})$).
The second metric is the test mean squared error (MSE).
We divided $[T^{\mathrm{c}}, T^{\mathrm{e}}]$ into 100 equal time-bins and compared the predicted number of events with ground truth values in each time-bins.
The time-bin size is 1.56 hours for the Bikeshare data, 39 minutes for the Taxi data, and 4.68 hours for the Crime data.
Detailed definitions are given in \cref{subsec:MSE}.

\textbf{Hyperparameters.}
We tuned hyperparameters of all methods by grid search on the validation set.
Specifically, 100 epochs were trained, and models were selected based on log-likelihood of the validation dataset for all hyperparameter combinations and all epochs.
In each method, the batch size was chosen from \{32, 64\}.
For the proposed method, the unit size in each layer of the MNN was chosen from \{128, 256, 512\} units, while the FNN had \{1, 2\} layers.
Each dataset is considered to have a periodicity of one day, thus we set $\tau=1$ day.
More details on the hyperparameters are described in \cref{sec:hyper}.

\begin{table}[t]
    \centering
    \setlength{\tabcolsep}{4pt}
    \footnotesize
    \begin{tabular}{lrrrrrr}
        \toprule
                   & \multicolumn{2}{c}{Bikeshare}                                                                                                                                                                                                             & \multicolumn{2}{c}{Taxi}                                               & \multicolumn{2}{c}{Crime}                                              \\ \cmidrule(lr){2-3}\cmidrule(lr){4-5}\cmidrule(lr){6-7}
                   & \multicolumn{1}{c}{NLL}                                                                                                                                                                                         & \multicolumn{1}{c}{MSE} & \multicolumn{1}{c}{NLL}                      & \multicolumn{1}{c}{MSE} & \multicolumn{1}{c}{NLL}                      & \multicolumn{1}{c}{MSE} \\ \midrule
        HPP        & 111.15                                                                                                                                                                                                          & 24.30                   & -347.0                                       & 23.12                   & 10.05                                        & 0.909                   \\
        NNIPP      & 134.89                                                                                                                                                                                                          & 24.14                   & -288.3                                       & 25.06                   & 5.18                                         & 0.722                   \\
        NM ($l=1$) & 108.46                                                                                                                                                                                                          & 23.17                   & -336.7                                       & 21.31                   & 3.49                                         & 0.705                   \\
        NM ($l=2$) & 62.01                                                                                                                                                                                                           & 19.44                   & -345.6                                       & 20.43                   & 3.06                                         & 0.700                   \\
        NM ($l=3$) & 105.19                                                                                                                                                                                                          & 21.25                   & -341.2                                       & 20.41                   & 2.80                                         & 0.694                   \\
        NM ($l=4$) & 91.17                                                                                                                                                                                                           & 21.38                   & -351.4                                       & 19.61                   & 2.43                                         & 0.688                   \\
        HARMLESS   & -\tablefootnote{HARMLESS is blank because it is not possible to calculate the appropriate NLL from the definition of the intensity function. Details are given in \cref{subsec:harmless}.\label{foot:harmless}} & 26.79                   & -\footnotemark[\getrefnumber{foot:harmless}] & 37.37                   & -\footnotemark[\getrefnumber{foot:harmless}] & 0.968                   \\
        Proposed   & \textbf{-14.31}                                                                                                                                                                                                 & \textbf{13.26}          & \textbf{-391.8}                              & \textbf{16.17}          & \textbf{0.44}                                & \textbf{0.666}          \\
        \bottomrule
    \end{tabular}
    \caption{Performance comparison (Lower is better). \label{tab:result}}
\end{table}
\subsection{Quantitative Results\label{subsec:results}}
\cref{tab:result} shows the prediction performance of the different methods in terms of the NLL and MSE on the three datasets.
The results indicate the superiority of our approach.
HARMLESS had the worst performance because it assumes Hawkes processes, but none of the dataset match this.
NNIPP had the next worst performance for Bikeshare and Taxi dataset because it did not use the support set of the target task, but instead produced the same prediction output for all tasks.
This suggests that each task is not always similar and should be treated as distinct tasks.
In fact, HPP using the support set of the target task, its intensity function fitted the task, and HPP performance was better than NNIPP for the Bikeshare and Taxi dataset.
NM performance was inferior to the proposed method, but showed the next best performance.
This indicates that combining inhomogeneous Poisson processes with meta-learning is appropriate for the three datasets.
When the number of inner loops are $l=3,4$ for Bikeshare dataset, NM showed lower performance than when $l=2$.
This result indicates that increasing $l$ does not necessarily improve the performance.

The proposed method takes the same or shorter time than NM ($l=1,2$) for both training and prediction.
The maximum GPU memory for train required by proposed method was only \SI{63}{\percent} of NM ($l=1$) when the hyperparameters are set to suppress memory.
From these results, the proposed method exhibits not only superior prediction performance but also demonstrates efficient computation.
More details are described in \cref{sec:computational}.
\begin{table}[t]
    \centering
    \setlength{\tabcolsep}{2pt}
    \footnotesize
    \begin{tabular}{lrrrrrr}
        \toprule
                           & \multicolumn{2}{c}{Bikeshare}                     & \multicolumn{2}{c}{Taxi}                          & \multicolumn{2}{c}{Crime}                         \\ \cmidrule(lr){2-3}\cmidrule(lr){4-5}\cmidrule(lr){6-7}
                           & \multicolumn{1}{c}{NLL} & \multicolumn{1}{c}{MSE} & \multicolumn{1}{c}{NLL} & \multicolumn{1}{c}{MSE} & \multicolumn{1}{c}{NLL} & \multicolumn{1}{c}{MSE} \\ \midrule
        Naive              & 54.39                   & 18.64                   & -360.2                  & 18.24                   & 6.02                    & 0.728                   \\
        w/o Periodicity    & 44.59                   & 18.64                   & -377.7                  & 16.88                   & 0.75                    & 0.669                   \\
        w/o Urban Contexts & -8.12                   & 13.45                   & -379.7                  & 16.61                   & 1.41                    & 0.672                   \\
        Full model         & \textbf{-14.31}         & \textbf{13.26}          & \textbf{-391.8}         & \textbf{16.17}          & \textbf{0.44}           & \textbf{0.666}          \\
        \bottomrule
    \end{tabular}
    \caption{Ablation study.\label{tab:ablation_result}
        \textmd{Naive excludes both factors.}
    }
\end{table}
\begin{figure*}[t]
    \centering
    \begin{minipage}[b]{.30\linewidth}
        \includegraphics[width=\linewidth]{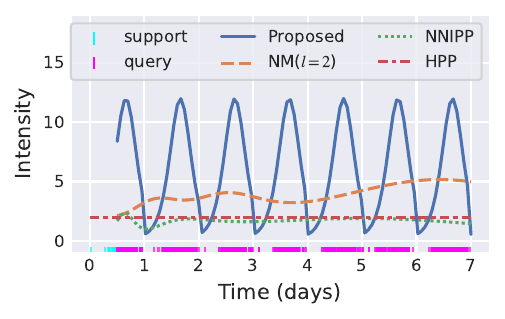}
        \subcaption{Bikeshare\label{fig:citibike_sample1}}
    \end{minipage}
    \begin{minipage}[b]{.30\linewidth}
        \includegraphics[width=\linewidth]{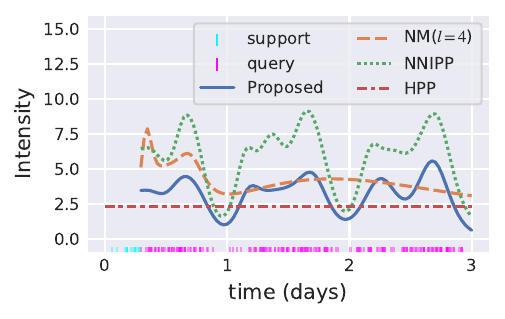}
        \subcaption{Taxi\label{fig:taxi_sample1}}
    \end{minipage}
    \begin{minipage}[b]{.30\linewidth}
        \includegraphics[width=\linewidth]{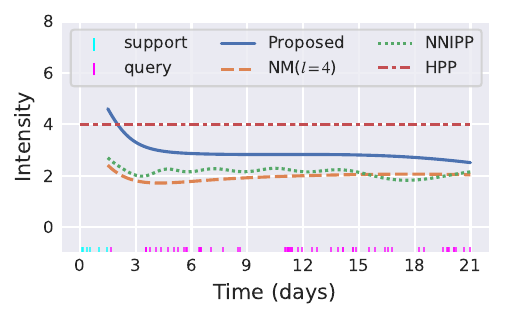}
        \subcaption{Crime\label{fig:crime_sample1}}
    \end{minipage}
    \caption{Examples of predicted intensities.\label{fig:result_example}
        \textmd{
            The intensity functions predicted by different methods.
            The cyan spikes at the bottom of the figure represent events in the support set, and the magenta spikes represent events in the query set.
        }
    }
\end{figure*}

\subsection{Ablation Analysis\label{subsec:ablation}}
To verify the effects of the external factors (i.e., temporal periodicity and urban contexts), we evaluated variants of the proposed method with and without these factors in \cref{tab:ablation_result}.
\cref{tab:ablation_result} shows that excluding each factor degrades performance, and excluding both of the factors (i.e., \emph{Naive}) further degrades it.
The results prove that the proposed method effectively incorporates these external factors.
Even \emph{Naive} performed better on the Bikeshare and Taxi datasets than the other methods.
This suggests that meta-learning with black-box adaptation, which does not require adaptation based on the gradient method, is promising for point processes.
The other ablation analysis is in \cref{sec:ablation_lstm}.

\subsection{Qualitative Results\label{subsec:result_analysis}}
\cref{fig:result_example} compares the intensity functions predicted by each method\footnote{HARMLESS was excluded due to performance issues.}.
Each line represents the intensity function of each method.
In \cref{fig:citibike_sample1}, the proposed method captured the periodic pattern.
In contrast, NM did not capture periodic patterns; its intensity function did not change significantly.
\cref{fig:taxi_sample1} shows that the proposed method captured the periodic pattern with two peaks (the morning and evening commuter rush hours).
In contrast, NM did not consider the periodic pattern and did not catch peaks after two days.
\cref{fig:crime_sample1} shows that the sparse event occurrence and no periodicity resulted in almost constant intensities for all methods.
The periodic term in the proposed method was zero, which captured the lack of periodicity in the data.

\begin{figure}[t]
    \begin{minipage}[b]{0.32\linewidth}
        \centering
        \includegraphics[width=\linewidth]{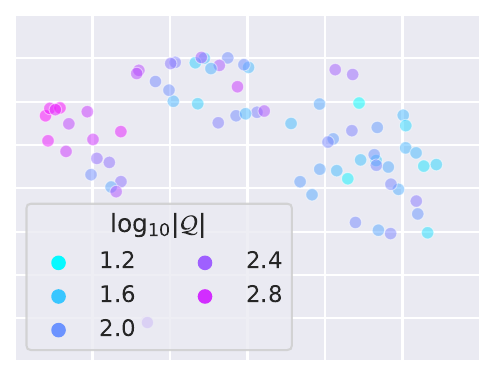}
        \subcaption{Bikeshare}
    \end{minipage}
    \begin{minipage}[b]{0.32\linewidth}
        \centering
        \includegraphics[width=\linewidth]{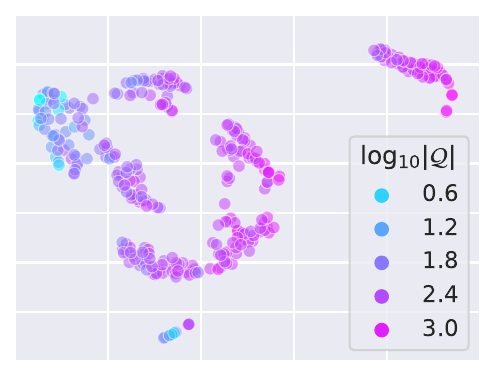}
        \subcaption{Taxi}
    \end{minipage}
    \begin{minipage}[b]{0.32\linewidth}
        \centering
        \includegraphics[width=\linewidth]{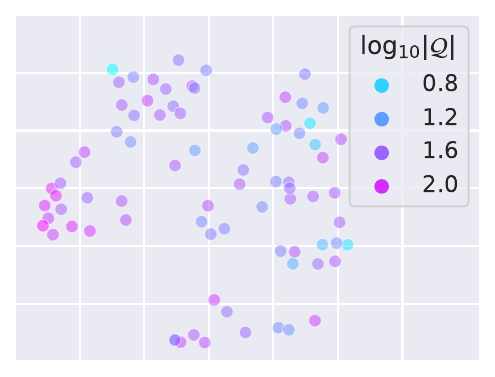}
        \subcaption{Crime}
    \end{minipage}
    \caption{Visualization of task representations $\bm{z}$ by $t$-SNE.\label{fig:tSNE}
        \textmd{
            The color denotes the number of events in the query set $\mathcal{Q}$.
        }
    }
\end{figure}
To confirm the effectiveness of our task representation encoder, we visualized the learned task representation $\bm{z}$ by using $t$-SNE \cite{van2014accelerating} in \cref{fig:tSNE}.
These figures show that the task representations reflected the number of events in the query set, even though it was calculated only from the support set and urban contexts.
Therefore, the proposed task representation encoder allows high prediction performance even for unknown short event sequences.

\begin{figure}[t]
    \centering
    \begin{minipage}[t]{.9\linewidth}
        \begin{minipage}[b]{\linewidth}
            \centering
            \begin{minipage}[b]{.22\linewidth}
                \includegraphics[width=\linewidth]{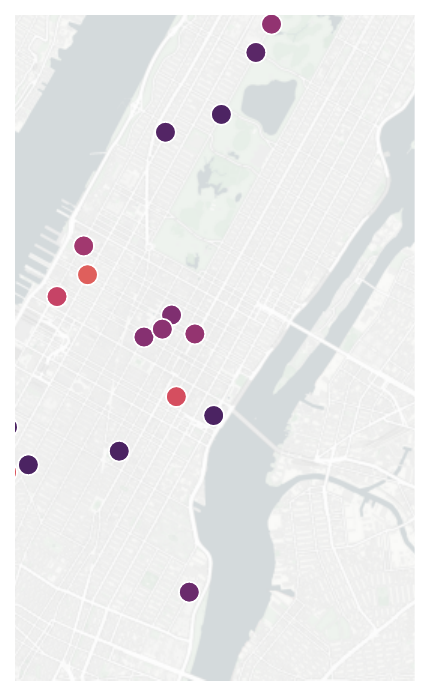}
                \subcaption*{Proposed}
            \end{minipage}
            \begin{minipage}[b]{.22\linewidth}
                \includegraphics[width=\linewidth]{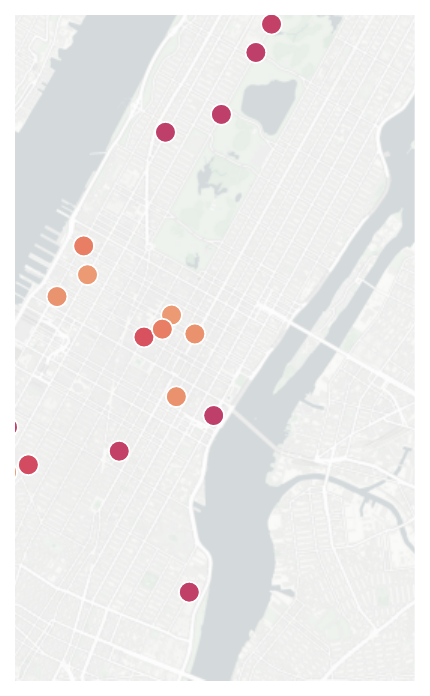}
                \subcaption*{NM}
            \end{minipage}
            \begin{minipage}[b]{.22\linewidth}
                \includegraphics[width=\linewidth]{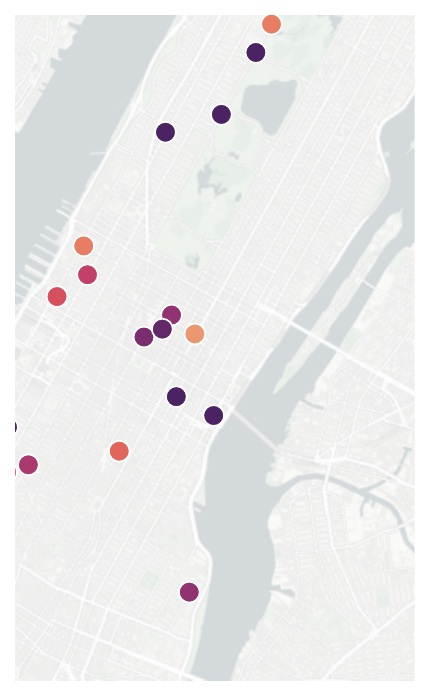}
                \subcaption*{Ground truth}
            \end{minipage}
            \subcaption{rush-hour ($t_0$ to 6-th day from 4:00 PM. to 5:00 PM.)\label{fig:heatmap_rush}}
        \end{minipage}
        \begin{minipage}[b]{\linewidth}
            \centering
            \begin{minipage}[b]{.22\linewidth}
                \includegraphics[width=\linewidth]{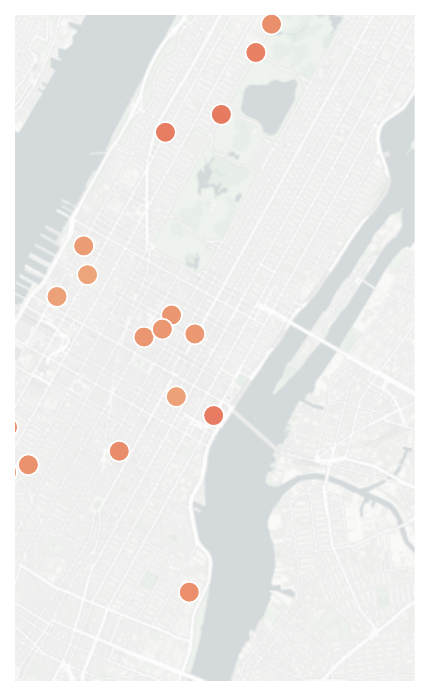}
                \subcaption*{Proposed}
            \end{minipage}
            \begin{minipage}[b]{.22\linewidth}
                \includegraphics[width=\linewidth]{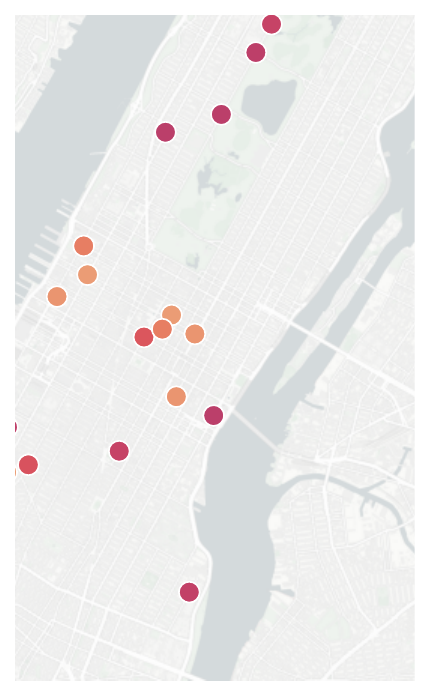}
                \subcaption*{NM}
            \end{minipage}
            \begin{minipage}[b]{.22\linewidth}
                \includegraphics[width=\linewidth]{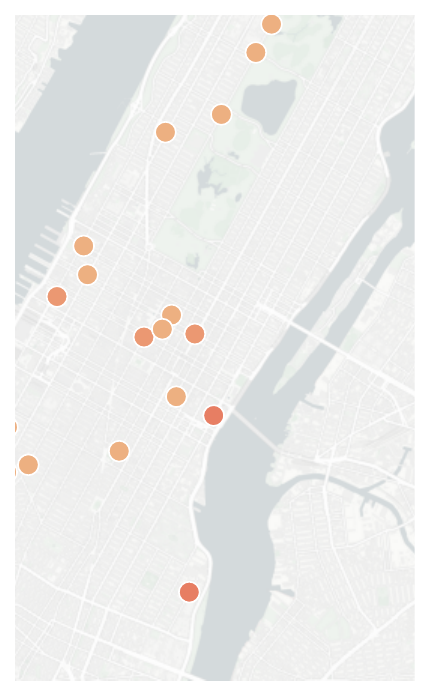}
                \subcaption*{Ground truth}
            \end{minipage}
            \subcaption{off-peak ($t_0$ to 6-th day from 4:00 AM. to 5:00 AM.)\label{fig:heatmap_off}}
        \end{minipage}

    \end{minipage}
    \begin{minipage}[p]{.08\linewidth}
        \includegraphics[width=\linewidth]{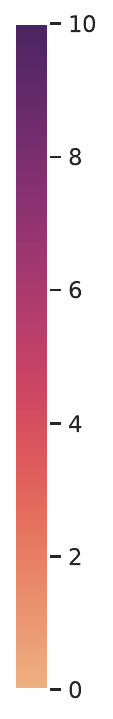}
    \end{minipage}
    \caption{Predicted number of pick-up events during two-time intervals for the bike stations in Manhattan (NYC).\label{fig:heatmap}}
\end{figure}
\cref{fig:heatmap} depicts the predicted numbers of pick-up events for the bike stations in Manhattan (NYC) during the two-time intervals.
The proposed method accurately predicted the changes in the pick-up demand at both rush-hour (\cref{fig:heatmap_rush}) and off-peak times (\cref{fig:heatmap_off}).
NMs, on the other hand, make similar predictions for both times.

\section{Conclusion\label{sec:end}}
In this paper, we proposed a new meta-learning approach for predicting future events given short sequences.
The proposed method uses a task representation encoder to allow short sequences to be embedded into task representations via RNN.
It models the intensity of the point process via MNNs through its input of the task representations.
We further extended the task representation to incorporate urban contexts and the MNN to incorporate temporal periodicity.
These factors are significant in human activities.
Experiments on multiple real-world datasets demonstrated that the proposed method has higher prediction performance than several existing methods.
Our future work includes replacing the RNN with another sequence model, such as a transformer, and addressing more complex features like images.

\bibliography{reference}

\appendix
\numberwithin{equation}{section}
\numberwithin{table}{section}
\numberwithin{figure}{section}
\numberwithin{algocf}{section}
\numberwithin{theorem}{section}
\section{Proof of Periodicity \label{sec:proof}}
\begin{theorem}\label{theorem:periodicity}
  $\lambda_\cycle(t;{\bm{z}}^{(m)})$, the derivative of $\Lambda_\cycle(t;{\bm{z}}^{(m)})$ in \cref{eq:cycle_dif_by_mnn}, is a periodic function with period $\tau$.
\end{theorem}
\begin{proof}
  From the definition and \cref{eq:cycle_dif_by_mnn},
  \begin{align}
    \MoveEqLeft \lambda_\cycle\left(t;{\bm{z}}^{(m)}\right)                             \nonumber                                                                          \\
     & = \frac{\partial \Lambda_\cycle\left(t;\bm{z}^{(m)}\right)}{\partial t} \nonumber                                                                                   \\
    \begin{split}
      & = \frac{\partial}{\partial t} s\left(f_\cycle\left(t',\bm{z}^{(m)}\right) - f_\cycle\left(0,\bm{z}^{(m)}\right)\right)
      \\
      & \quad + \frac{\partial}{\partial t} s\left\lfloor\frac{t}{\tau}\right\rfloor \left( f_\cycle\left(\tau,\bm{z}^{(m)}\right) - f_\cycle\left(0,\bm{z}^{(m)}\right) \right),\nonumber
    \end{split} \\
     & = \frac{\partial}{\partial t} sf_\cycle\left(t',\bm{z}^{(m)}\right)
    + \frac{\partial}{\partial t} s\left\lfloor\frac{t}{\tau}\right\rfloor  f_\cycle\left(\tau,\bm{z}^{(m)}\right)  ,\label{eq:lambda}
  \end{align}
  and
  \begin{align}
    \MoveEqLeft \lambda_\cycle\left(t+\tau;{\bm{z}}^{(m)}\right)                             \nonumber                                               \\
     & = \frac{\partial \Lambda_\cycle\left(t+\tau;\bm{z}^{(m)}\right)}{\partial t} \nonumber                                                        \\
    \begin{split}
      & = \frac{\partial}{\partial t} sf_\cycle\left(t + \tau -  \tau \left\lfloor\frac{t+\tau}{\tau}\right\rfloor,\bm{z}^{(m)}\right)  \\
      & \quad + \frac{\partial}{\partial t} s\left\lfloor\frac{t+\tau}{\tau}\right\rfloor  f_\cycle\left(\tau,\bm{z}^{(m)}\right)   \nonumber
    \end{split} \\
    \begin{split}
      & = \frac{\partial}{\partial t} sf_\cycle\left(t -  \tau \left\lfloor\frac{t}{\tau}\right\rfloor,\bm{z}^{(m)}\right)  \\
      & \quad + \frac{\partial}{\partial t} s\left(\left\lfloor\frac{t}{\tau}\right\rfloor  +1\right)f_\cycle\left(\tau,\bm{z}^{(m)}\right)   \label{eq:lambda_add_tau}
    \end{split}
  \end{align}
  From \cref{eq:lambda,eq:lambda_add_tau},
  \begin{align}
    \MoveEqLeft \lambda_\cycle\left(t+\tau;{\bm{z}}^{(m)}\right) - \lambda_\cycle\left(t;{\bm{z}}^{(m)}\right) \nonumber \\
     & = \frac{\partial}{\partial t} s  f_\cycle\left(\tau,\bm{z}^{(m)}\right) \nonumber                                 \\
     & = 0.
  \end{align}
  Thus,  $\lambda_\cycle\left(t;{\bm{z}}^{(m)}\right) = \lambda_\cycle\left(t + \tau;{\bm{z}}^{(m)}\right)$.
\end{proof}

\section{Dataset Statistics and List of Urban Contexts\label{sec:dataset}}
\cref{tab:statistics} shows the statistics of event data in each dataset and the mean (M) and standard deviation (SD) number of events in each event dataset.
\cref{tab:urban} shows the list of urban contexts, which were downloaded from Community District Profiles\footnote{\url{https://communityprofiles.planning.nyc.gov/}}.
\begin{table}[t]
  \centering
  \small
  \begin{tabular}{llrD{,}{\,\pm\,}{5}D{,}{\,\pm\,}{5}}
    \toprule
    \multicolumn{2}{c}{dataset}        & \multicolumn{1}{c}{$\abs{D}$} & \multicolumn{1}{c}{$\abs{\mathcal{S}}$ (M$\pm$SD)} & \multicolumn{1}{c}{$\abs{\mathcal{Q}}$ (M$\pm$SD)} \\ \midrule
    \multirow{3}{*}{Bikeshare} & train & 539                           & 10.5,13.5                                          & 263.5,295.1                                        \\
                               & val.  & 52                            & 6.3, 5.4                                           & 256.2,253.7                                        \\
                               & test  & 76                            & 8.0,10.2                                           & 240.9,305.6                                        \\ \midrule
    \multirow{3}{*}{Taxi}      & train & 970                           & 33.6,37.4                                          & 346.9,346.2                                        \\
                               & val.  & 276                           & 35.1,38.4                                          & 347.8,343.7                                        \\
                               & test  & 328                           & 31.6,30.8                                          & 364.2,350.8                                        \\ \midrule
    \multirow{3}{*}{Crime}     & train & 313                           & 6.3,2.0                                            & 39.8,28.0                                          \\
                               & val.  & 83                            & 6.1,1.7                                            & 43.4,29.9                                          \\
                               & test  & 80                            & 6.6,2.8                                            & 48.2,34.8                                          \\
    \bottomrule
  \end{tabular}
  \caption{Dataset statistics\label{tab:statistics}}

\end{table}
\begin{table}[t]
  \centering
  \begin{tabular}{ll}
    \toprule
    Category         & Item                              \\ \midrule
    Land Use         & Multifamily Walk-up               \\
    Land Use         & Industrial \& Manufacturing       \\
    Land Use         & 1 \& 2 Family                     \\
    Land Use         & Public Facilities \& Institutions \\
    Land Use         & Multifamily Elevator              \\
    Land Use         & Commercial \& Office              \\
    Land Use         & Open Space \& Outdoor Recreation  \\
    Land Use         & Vacant Land                       \\
    Land Use         & Mixed Res. \& Commercial          \\
    Land Use         & Transportation \& Utility         \\
    Land Use         & Parking Facilities                \\
    Community Assets & Public Schools                    \\
    Community Assets & Public Libraries                  \\
    Community Assets & Hospitals and Clinics             \\
    Community Assets & Parks                             \\
    Indicators       & Crime                             \\
    Indicators       & Street Cleanliness                \\
    Indicators       & Mean Commute to Work              \\ \bottomrule
  \end{tabular}
  \caption{List of urban contexts\label{tab:urban}}
\end{table}

\section{Comparison Methods\label{subsec:VSmethods_describe}}
\subsection{HPP\label{subsec:HPP}}
HPP \cite{pointprocess2005} is a model with a constant intensity function.
Here, the intensity function is determined based on the support set of each task.
That is, $\lambda(t) = {\abs{\mathcal{S}^{(m^*)}}}/{T^\mathrm{c}}$.

\subsection{NNIPP\label{subsec:NNIPP}}
NNIPP is a simple neural-network-based inhomogeneous Poisson process model without meta-learning.
As an inhomogeneous Poisson process \cite{pointprocess2005}, its intensity function varies with time.
The neural network used in NNIPP is the same as the MNN used in the proposed method's aperiodic term (except that the input is just time $t$).
That is, it is the proposed method but without RNN and the periodic term.
Its training and prediction processes consider all event data as the same task.
The intensity function and loss used in training are similar to \cref{eq:noncycle_dif_by_mnn,eq:loss}, where modules involved in $\bm{z}$ are removed.

\subsection{NM\label{subsec:maml}}
NM is a model that adapts MAML \cite{finn2017model} to NNIPP.
That is, NM models it with a neural network identical to NNIPP.
MAML performs double-loop training to achieve meta-learning.
In the inner loop, parameters are updated several times using back-propagation to adapt to the new task.
The number of inner loops, denoted as $l$ in this paper, was varied from 1 to 4.
The inner loop objective function is given by,
\begin{align}
  \hat{\theta}'         & = \arg \min_{\theta} E_{\mathcal{S} \sim \mathcal{D}}[L_\theta(\mathcal{S})] \label{eq:object_inner} \\
  L_\theta(\mathcal{S}) & = -\sum_{t \in \mathcal{S}}\log \lambda_\theta(t)   + \Lambda_{\theta}(T^{\mathrm{c}})
  - \Lambda_{\theta}(0) \label{eq:loss_inner_maml}.
\end{align}
The outer loop objective function is given by
\begin{align}
  \hat{\theta}                                  & = \arg \min_{\theta} E_{(\mathcal{S},\mathcal{Q}) \sim \mathcal{D}}[L_{\hat{\theta}'}(\mathcal{S}\cup\mathcal{Q})] \label{eq:object_outer} \\
  L_{\hat{\theta}'}(\mathcal{S}\cup\mathcal{Q}) & =
  -\sum_{t \in  \mathcal{S}\cup\mathcal{Q}}\log \lambda_{\hat{\theta}'}(t)
  + \Lambda_{\hat{\theta}'}(T^{\mathrm{e}})
  - \Lambda_{\hat{\theta}'}(0)
  \label{eq:loss_outer_maml}.
\end{align}
\begin{algorithm}[t]
  \DontPrintSemicolon
  \KwIn{Event datasets $\mathcal{D}$, query time interval $[T^\mathrm{c},T^\mathrm{e}]$, number of inner loop $l$}
  \KwOut{Trained neural network parameters $\theta$}
  Determine $s$ from $\mathcal{D}$\;
  Initialize general model parameter $\theta$\;
  \While{\textrm{not done}}{
    Randomly sample task $m$ from $M$\;
    Divide $X^{(m)}_{T^\mathrm{e}}$ into support set $\mathcal{S}$ and query set $\mathcal{Q}$\;
    Set task specific model parameters $\theta' \leftarrow \theta$\;
    \Repeat{$l$ times}{
      Calculate loss using \cref{eq:object_inner,eq:loss_inner_maml}\;
      Update parameters $\theta'$ using loss and  gradients
    }
    Calculate loss by a version of \cref{eq:object_outer,eq:loss_outer_maml}, and its second derivative\;
    Update general model parameters, $\theta$, using the loss and its second derivative
  }

  \caption{Training procedure of NM\label{algo:training_maml}}
\end{algorithm}

\subsection{HARMLESS\label{subsec:harmless}}
HARMLESS is a model that adapts MAML \cite{finn2017model} to Hawkes process.
The intensity function is given by
\begin{equation}
  \lambda(t \mid \mathcal{X}_t) = \mu + \sum_{k=1}^{N^\mathrm{k}}\sum_{t_i\in \mathcal{X}_t} \delta_k\omega_k e^{-\omega_k(t-t_i)},
\end{equation}
where $N^\mathrm{k}$ is the number of kernel functions, and $\mu$, $\delta$, and $\omega$ are the learning parameters.
Note that $\mathcal{X}_t$ is similar to $X_t$ but does not include just $t$, i.e., $\mathcal{X}_t=\{t_1,\dots,t_N\},(0\leq t_1\leq \dots \leq t_N < t)$.
The NLL is not computable, as the intensity function requires information on all events that have occurred before that time, $t$, including query set $\mathcal{Q}$.
For the same reason, the MSE values are the results of simulations using the thinning method \cite{ogata1981lewis}.

\subsection{Proposed Method\label{subsec:proposed_setting}}
As the neural networks in our model, we used bi-directional LSTM \cite{sak2014long} for RNN and fully-connected neural networks for MNN.
We set $K_{\mathrm{z}} = K_{\mathcal{S}} = 128$.
LSTM was designed with 128 units in one layer.
FNN was structured with 128 units in one layer, and the activation functions used were hyperbolic tangent.
For the MNNs, all weights were restricted to be non-negative, and the activation functions used were hyperbolic tangent in the hidden layers and softplus in the output layer, as in previous study \cite{omi2019fully}.
Based on the results of preliminary experiments, the initial values of the weights of all MNNs were set to follow glorot uniform \cite{glorot2010understanding}.
Note that LSTM inputs are just the events in the support set, i.e., past events, even in the test phase.
Therefore, using bi-directional LSTM does not invalidate the proposal.
\cref{sec:ablation_lstm} provides a comparison with uni-directional LSTM.

\section{Experimental Setup Details\label{sec:setup_detail}}
\subsection{Implementation and Environment}
We used python3.9.5 and pytorch1.10.2 \cite{paszke2019pytorch} to implement the algorithms.
Each experiment was run on a pair of Xeon Platinum 8176 (2.10GHz) and a NVIDIA TITAN V.
We also used higher (version 0.2.1) \cite{grefenstette2019generalized} to achieve MAML adaptation.

\subsection{Optimization}
For HARMLESS, NNIPP, NM and proposed method, we used Adam \cite{kingma2014adam} with ($\alpha=0.001,\beta_1=0.9,\beta_2=0.999,\epsilon=10^{-8}$) for optimization.
Note that HPP does not require an optimization algorithm such as Adam.

\subsection{Metrics' Definitions\label{subsec:MSE}}
NLL is a version of \cref{eq:loss} with only the query set as the test data,
\begin{align}\label{eq:nll}
  L_\theta\left(\mathcal{S}^{(m^*)},\mathcal{Q}^{(m^*)}\right)
   & = -\sum_{t \in \mathcal{Q}^{(m^*)}}
  \log \frac{\partial \Lambda_\theta\left(t;\bm{z}^{(m^*)}\right)}{\partial t} \nonumber \\
   & + \Lambda_\theta\left(T^{\mathrm{e}};\bm{z}^{(m^*)}\right)
  - \Lambda_\theta\left(T^{\mathrm{c}};\bm{z}^{(m^*)}\right).
\end{align}

We calculated the MSE by dividing the continuous time into several intervals.
Specifically, we divided $[T^{\mathrm{c}}, T^{\mathrm{e}}]$ and compared the predicted number of events with the ground truth in each interval $\{[T_j^{\mathrm{a}},T_j^\mathrm{b}]\}_{k=j}^J$.
The MSE is defined as
\begin{equation}
  \mathrm{MSE}=\frac{1}{J}\sum_{j=1}^I \left[\abs{X^{(m^*)}_{T_j^\mathrm{b}}} - \abs{X^{(m^*)}_{T_j^\mathrm{a}}} - \int_{T_j^\mathrm{a}}^{T_j^\mathrm{b}} \hat{\lambda}(u)du\right]^2.
\end{equation}
In this paper, we set $J=100$.

\section{Hyperparameters\label{sec:hyper}}
We tuned hyperparameters of all methods by grid search on the validation set.
We targeted five hyperparameters for our grid search.
The other hyperparameters were set as described in \cref{subsec:VSmethods_describe,sec:setup_detail}.
The search range for each target hyperparameter is shown in \cref{tab:hyper_range}.
Note that the number of units for the proposed method with the periodic term only is half that of the other methods because they use double the number of MNN layers, i.e., periodic and aperiodic terms.
The number of units for uni-directional LSTM is determined to keep the number of parameters or units per layer the same as for bi-directional LSTM i.e., proposed method.
Also, the MAML's weight decay may apply different values for the inner and outer loops.
The final hyperparameters selected for each method are shown in \cref{tab:hyper_bike,tab:hyper_taxi,tab:hyper_crime}.

\begin{table*}
  \centering
  \begin{tabular}[t]{ll}
    \toprule
    hyperparameter                                                               & range                \\ \midrule
    batch size                                                                   & $\{ 32, 64\}$        \\
    weight decay                                                                 & $\{0.0, 0.0005\}$    \\
    the number of units in each layer of the MNN (methods with periodic term)    & $\{128, 256, 512\}$  \\
    the number of units in each layer of the MNN (methods without periodic term) & $\{256, 512, 1024\}$ \\
    the number of layers of the FNN                                              & $\{1, 2\}$           \\
    the number of units in layer of the LSTM (uni-directional LSTM)              & $\{128, 256\}$       \\
    the number of kernel functions (HARMLESS)                                    & $\{1,2,3,4\}$        \\
    the number of inner loops (HARMLESS and NM)                                  & $\{1,2,3,4\}$        \\
    \bottomrule
  \end{tabular}
  \caption{Search range for each hyperparameter. \label{tab:hyper_range}}
\end{table*}

\begin{table*}
  \centering

  \begin{tabular}[t]{lrrrrrrrr}
    \toprule
                       & \multicolumn{1}{c}{\multirow{2}{*}{batch size}} & \multicolumn{1}{c}{MNN} & \multicolumn{1}{c}{FNN} & \multicolumn{1}{c}{RNN} & \multicolumn{1}{c}{Weight } & \multicolumn{1}{c}{Weight Decay}   & \multicolumn{1}{c}{\# of kernel} & \multicolumn{1}{c}{\# of }      \\
                       &                                                 & \# of units             & \# of layers            & \# of units             & \multicolumn{1}{c}{Decay}   & \multicolumn{1}{c}{for inner loop} & \multicolumn{1}{c}{functions}    & \multicolumn{1}{c}{inner loops} \\ \midrule
    HARMLESS           & 32                                              & -                       & -                       & -                       & 0.0                         & 0.0                                & 4                                & 2                               \\
    NM ($l=1$)         & 32                                              & 1024                    & -                       & -                       & 0.0                         & 0.0005                             & -                                & 1                               \\
    NM ($l=2$)         & 32                                              & 256                     & -                       & -                       & 0.0005                      & 0.0005                             & -                                & 2                               \\
    NM ($l=3$)         & 32                                              & 256                     & -                       & -                       & 0.0                         & 0.0005                             & -                                & 3                               \\
    NM ($l=4$)         & 32                                              & 512                     & -                       & -                       & 0.0                         & 0.0                                & -                                & 4                               \\
    NNIPP              & 32                                              & 1024                    & -                       & -                       & 0.0005                      & -                                  & -                                & -                               \\
    Proposed           & 32                                              & 512                     & 2                       & -                       & 0.0005                      & -                                  & -                                & -                               \\\midrule
    Naive              & 32                                              & 256                     & -                       & -                       & 0.0005                      & -                                  & -                                & -                               \\
    w/o Periodicity    & 32                                              & 1024                    & 1                       & -                       & 0.0005                      & -                                  & -                                & -                               \\
    w/o Urban Contexts & 32                                              & 512                     & -                       & -                       & 0.0                         & -                                  & -                                & -                               \\\midrule
    uni-directional    & 32                                              & 128                     & 1                       & 128                     & 0.0005                      & -                                  & -                                & -                               \\
    \bottomrule
  \end{tabular}
  \caption{Selected hyperparameters for Bikeshare dataset. \label{tab:hyper_bike}}
\end{table*}

\begin{table*}
  \centering

  \begin{tabular}[t]{lrrrrrrrr}
    \toprule
                       & \multicolumn{1}{c}{\multirow{2}{*}{batch size}} & \multicolumn{1}{c}{MNN} & \multicolumn{1}{c}{FNN} & \multicolumn{1}{c}{RNN} & \multicolumn{1}{c}{Weight } & \multicolumn{1}{c}{Weight Decay}   & \multicolumn{1}{c}{\# of kernel} & \multicolumn{1}{c}{\# of }      \\
                       &                                                 & \# of units             & \# of layers            & \# of units             & \multicolumn{1}{c}{Decay}   & \multicolumn{1}{c}{for inner loop} & \multicolumn{1}{c}{functions}    & \multicolumn{1}{c}{inner loops} \\ \midrule
    HARMLESS           & 32                                              & -                       & -                       & -                       & 0.0                         & 0.0                                & 4                                & 1                               \\
    NM ($l=1$)         & 64                                              & 512                     & -                       & -                       & 0.0005                      & 0.0                                & -                                & 1                               \\
    NM ($l=2$)         & 32                                              & 1024                    & -                       & -                       & 0.0005                      & 0.0005                             & -                                & 2                               \\
    NM ($l=3$)         & 32                                              & 256                     & -                       & -                       & 0.0005                      & 0.0                                & -                                & 3                               \\
    NM ($l=4$)         & 32                                              & 512                     & -                       & -                       & 0.0                         & 0.0                                & -                                & 4                               \\
    NNIPP              & 64                                              & 1024                    & -                       & -                       & 0.0005                      & -                                  & -                                & -                               \\
    Proposed           & 32                                              & 128                     & 2                       & -                       & 0.0                         & -                                  & -                                & -                               \\\midrule
    Naive              & 64                                              & 256                     & -                       & -                       & 0.0005                      & -                                  & -                                & -                               \\
    w/o Periodicity    & 64                                              & 512                     & 2                       & -                       & 0.0005                      & -                                  & -                                & -                               \\
    w/o Urban Contexts & 64                                              & 512                     & -                       & -                       & 0.0                         & -                                  & -                                & -                               \\\midrule
    uni-directional    & 64                                              & 128                     & 2                       & 128                     & 0.0005                      & -                                  & -                                & -                               \\
    \bottomrule
  \end{tabular}
  \caption{Selected hyperparameters for Taxi dataset. \label{tab:hyper_taxi}}
\end{table*}

\begin{table*}
  \centering

  \begin{tabular}[t]{lrrrrrrrr}
    \toprule
                       & \multicolumn{1}{c}{\multirow{2}{*}{batch size}} & \multicolumn{1}{c}{MNN} & \multicolumn{1}{c}{FNN} & \multicolumn{1}{c}{RNN} & \multicolumn{1}{c}{Weight } & \multicolumn{1}{c}{Weight Decay}   & \multicolumn{1}{c}{\# of kernel} & \multicolumn{1}{c}{\# of }      \\
                       &                                                 & \# of units             & \# of layers            & \# of units             & \multicolumn{1}{c}{Decay}   & \multicolumn{1}{c}{for inner loop} & \multicolumn{1}{c}{functions}    & \multicolumn{1}{c}{inner loops} \\ \midrule
    HARMLESS           & 32                                              & -                       & -                       & -                       & 0.0005                      & 0.0005                             & 2                                & 2                               \\
    NM ($l=1$)         & 64                                              & 256                     & -                       & -                       & 0.0                         & 0.0005                             & -                                & 1                               \\
    NM ($l=2$)         & 64                                              & 256                     & -                       & -                       & 0.0005                      & 0.0                                & -                                & 2                               \\
    NM ($l=3$)         & 64                                              & 256                     & -                       & -                       & 0.0                         & 0.0005                             & -                                & 3                               \\
    NM ($l=4$)         & 32                                              & 256                     & -                       & -                       & 0.0                         & 0.0005                             & -                                & 4                               \\
    NNIPP              & 64                                              & 512                     & -                       & -                       & 0.0                         & -                                  & -                                & -                               \\
    Proposed           & 64                                              & 512                     & 2                       & -                       & 0.0                         & -                                  & -                                & -                               \\\midrule
    Naive              & 64                                              & 256                     & -                       & -                       & 0.0005                      & -                                  & -                                & -                               \\
    w/o Periodicity    & 32                                              & 512                     & 2                       & -                       & 0.0                         & -                                  & -                                & -                               \\
    w/o Urban Contexts & 32                                              & 128                     & -                       & -                       & 0.0005                      & -                                  & -                                & -                               \\\midrule
    uni-directional    & 64                                              & 128                     & 2                       & 256                     & 0.0005                      & -                                  & -                                & -                               \\
    \bottomrule
  \end{tabular}
  \caption{Selected hyperparameters for Crime dataset. \label{tab:hyper_crime}}
\end{table*}

\section{Computational Times and Resources\label{sec:computational}}
The proposed method's and NM's measured prediction times are shown in \cref{tab:time}.
The execution environment is as described in \cref{sec:setup_detail}.
The proposed method takes the same or shorter time than NM ($l=1,2$) for both training and prediction.
This can be attributed to the fact that the proposed method can be computed efficiently as it avoids adaptation based on the gradient method.

The maximum GPU memory for train and calculating NLL required by each method when the hyperparameters are set to suppress memory used is shown in \cref{tab:memory}.
The proposed method can be trained with an increase in GPU memory used of \SI{21}{\percent} to \SI{32}{\percent} from NNIPP.
This is because the proposed method achieves meta-learning only by adding network components such as RNNs.
NM requires over twice as much GPU memory, even with $l=1$.
This is because NM performs adaptation using the gradient method, which increases the size of the computational graph that must be kept.
Meanwhile, NM tended to use less memory for tests.
This is because memory could be reused for each task, as multiple tasks were not adapted at once due to the implementation of higher\cite{grefenstette2019generalized}
\begin{table}[t]
  \centering

  \small
  \begin{tabular}{clD{.}{.}{2}D{.}{.}{2}}
    \toprule
    dataset                    & \multicolumn{1}{c}{Method} & \multicolumn{1}{c}{Train ($\si{s}$)} & \multicolumn{1}{c}{Prediction ($\si{s}$)} \\ \midrule
    \multirow{6}{*}{Bikeshare} & NM ($l=1$)                 & 24.86                                & 2.56                                      \\
                               & NM ($l=2$)                 & 37.62                                & 3.52                                      \\
                               & NM ($l=3$)                 & 48.50                                & 4.26                                      \\
                               & NM ($l=4$)                 & 62.17                                & 5.65                                      \\
                               & Proposed                   & 27.72                                & 2.16                                      \\\midrule
    \multirow{6}{*}{Taxi}      & NM ($l=1$)                 & 49.55                                & 10.23                                     \\
                               & NM ($l=2$)                 & 79.23                                & 13.85                                     \\
                               & NM ($l=3$)                 & 97.19                                & 16.82                                     \\
                               & NM ($l=4$)                 & 126.40                               & 21.06                                     \\
                               & Proposed                   & 78.40                                & 10.94                                     \\\midrule
    \multirow{6}{*}{Crime}     & NM ($l=1$)                 & 14.39                                & 2.65                                      \\
                               & NM ($l=2$)                 & 21.54                                & 3.28                                      \\
                               & NM ($l=3$)                 & 29.14                                & 4.24                                      \\
                               & NM ($l=4$)                 & 37.13                                & 5.48                                      \\
                               & Proposed                   & 4.88                                 & 1.38                                      \\
    \bottomrule
  \end{tabular}
  \caption{Time taken for training and prediction per epoch. \label{tab:time}}
\end{table}
\begin{table*}
  \centering
  \begin{tabular}{lrrrrrr}
    \toprule
               & \multicolumn{2}{c}{Bikeshare}                        & \multicolumn{2}{c}{Taxi}                             & \multicolumn{2}{c}{Crime}                            \\ \cmidrule(lr){2-3}\cmidrule(lr){4-5}\cmidrule(lr){6-7}
               & \multicolumn{1}{c}{train} & \multicolumn{1}{c}{test} & \multicolumn{1}{c}{train} & \multicolumn{1}{c}{test} & \multicolumn{1}{c}{train} & \multicolumn{1}{c}{test} \\ \midrule
    NNIPP      & 0.168                     & 0.054                    & 0.238                     & 0.093                    & 0.025                     & 0.011                    \\
    NM ($l=1$) & 0.378                     & 0.044                    & 0.477                     & 0.047                    & 0.253                     & 0.042                    \\
    NM ($l=2$) & 0.566                     & 0.043                    & 0.739                     & 0.046                    & 0.443                     & 0.042                    \\
    NM ($l=3$) & 0.778                     & 0.043                    & 0.978                     & 0.057                    & 0.634                     & 0.043                    \\
    NM ($l=4$) & 0.987                     & 0.043                    & 1.267                     & 0.049                    & 0.824                     & 0.043                    \\
    Proposed   & 0.203                     & 0.072                    & 0.301                     & 0.122                    & 0.033                     & 0.019                    \\\bottomrule
  \end{tabular}
  \caption{Maximum GPU memory usage. All values are expressed in Gibibytes (\si{\gibi\byte}).\label{tab:memory}}
\end{table*}

\section{Ablation Analysis for LSTM\label{sec:ablation_lstm}}
\begin{table*}[t]
  \centering
  \small
  \begin{tabular}{lrrrrrr}
    \toprule
    \multicolumn{1}{c}{\multirow{2}{*}{LSTM}} & \multicolumn{2}{c}{Bikeshare}                     & \multicolumn{2}{c}{Taxi}                          & \multicolumn{2}{c}{Crime}                         \\ \cmidrule(lr){2-3}\cmidrule(lr){4-5}\cmidrule(lr){6-7}
                                              & \multicolumn{1}{c}{NLL} & \multicolumn{1}{c}{MSE} & \multicolumn{1}{c}{NLL} & \multicolumn{1}{c}{MSE} & \multicolumn{1}{c}{NLL} & \multicolumn{1}{c}{MSE} \\ \midrule

    uni-directional                           & 4.86                    & 15.26                   & -370.4                  & 18.23                   & 3.53                    & 0.706                   \\
    bi-directional                            & \textbf{-14.31}         & \textbf{13.26}          & \textbf{-391.8}         & \textbf{16.17}          & \textbf{0.44}           & \textbf{0.666}          \\
    \bottomrule
  \end{tabular}
  \caption{Performance comparison for LSTM.\label{tab:ablation_lstm}}
\end{table*}
To verify the characteristics of the proposed method, we compare it with bi-directional and uni-directional LSTM.
\cref{tab:ablation_lstm} shows that uni-directional LSTM has lower performance than bi-directional LSTM but better performance than the baseline methods for Bikeshare dataset and Taxi dataset (See~\cref{tab:result}).
This confirms that our framework does not rely on bi-directional LSTM and that bi-directional LSTM is useful for embedding the support set to obtain support representation.

\section{Qualitative Results for Urban Contexts}
\begin{figure*}[t]
  \begin{minipage}[b]{.32\linewidth}
    \includegraphics[width=\linewidth]{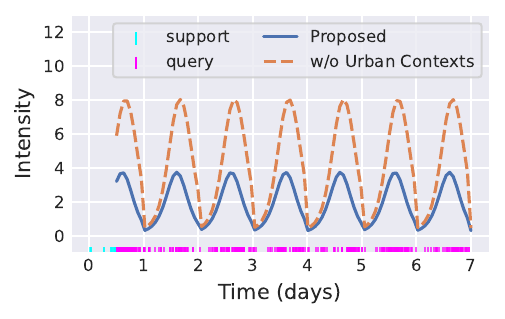}
    \subcaption{Bikeshare\label{fig:citibike_sample2}}
  \end{minipage}
  \begin{minipage}[b]{.32\linewidth}
    \includegraphics[width=\linewidth]{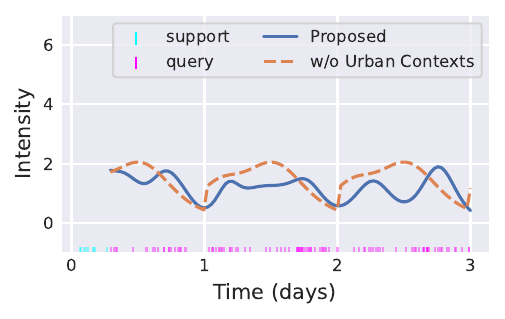}
    \subcaption{Taxi\label{fig:taxi_sample2}}
  \end{minipage}
  \begin{minipage}[b]{.32\linewidth}
    \includegraphics[width=\linewidth]{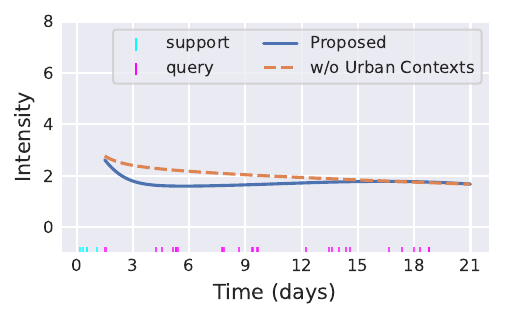}
    \subcaption{Crime\label{fig:crime_sample2}}
  \end{minipage}
  \caption{Differences with and without urban contexts.\label{fig:result_example2}}
\end{figure*}
\cref{fig:result_example2} compares the intensity functions predicted by Proposed and \emph{w/o Urban Contexts} to confirm the effect of urban contexts.
The urban contexts are significantly adjusted for the number of events for the Bikeshare dataset, as shown in \cref{fig:citibike_sample1}.
Indeed, in the urban contexts, the area of apartment blocks with elevators (multifamily elevator of land use) and mean commute time to work were correlated with the number of events in the query set, $r=0.647,p<0.001$ and $r=-0.548 ,p<0.001$, respectively.
As shown in \cref{fig:taxi_sample2}, the urban contexts contribute to the extraction of two peaks per day for the Taxi dataset.
No large differences by urban contexts could be observed for the Crime data, with only some adjustments for the number of events and other factors, as shown in \cref{fig:crime_sample2}.
The above shows that the urban contexts are beneficial in all datasets, although their utilities and effect sizes are different.

\end{document}